\documentclass{article}

\usepackage{arxiv}

\usepackage[utf8]{inputenc} 
\usepackage[T1]{fontenc}    
\usepackage[pagebackref,breaklinks,colorlinks]{hyperref}
\usepackage{url}            
\usepackage{booktabs}       
\usepackage{amsfonts}       
\usepackage{nicefrac}       
\usepackage{microtype}      
\usepackage{lipsum}

\usepackage{amssymb}
\usepackage{amsmath}
\usepackage[normalem]{ulem}
\usepackage{multirow}
\usepackage{array}
\usepackage{graphicx}
\usepackage{bm}
\usepackage{caption,subcaption}        
\usepackage{pifont}
\usepackage{xcolor}                    

\useunder{\uline}{\ul}{}

\title{Hedging Is Not All You Need: A Simple Baseline for Online Learning Under Haphazard Inputs}

\author{
    Himanshu Buckchash \\
    UiT The Arctic University of Norway \\
    Tromsø, Norway \\
    \texttt{himanshu.buckchash@uit.no} \\
    \And
    Momojit Biswas \\
    Jadavpur University \\
    Kolkata, India \\
    \texttt{mb16biswas@gmail.com} \\
    \And
    Rohit Agarwal \\
    UiT The Arctic University of Norway \\
    Tromsø, Norway \\
    \texttt{rohit.agarwal@uit.no} \\
    \And
    Dilip K. Prasad \\
    UiT The Arctic University of Norway \\
    Tromsø, Norway \\
    \texttt{dilip.prasad@uit.no} \\
}

\begin{document}
\maketitle

\begin{abstract}
    Handling haphazard streaming data, such as data from edge devices, presents a challenging problem. Over time, the incoming data becomes inconsistent, with missing, faulty, or new inputs reappearing. Therefore, it requires models that are reliable. Recent methods to solve this problem depend on a hedging-based solution and require specialized elements like auxiliary dropouts, forked architectures, and intricate network design. We observed that hedging can be reduced to a special case of weighted residual connection; this motivated us to approximate it with plain self-attention. In this work, we propose HapNet, a simple baseline that is scalable, does not require online backpropagation, and is adaptable to varying input types. All present methods are restricted to scaling with a fixed window; however, we introduce a more complex problem of scaling with a variable window where the data becomes positionally uncorrelated, and cannot be addressed by present methods. We demonstrate that a variant of the proposed approach can work even for this complex scenario. We extensively evaluated the proposed approach on five benchmarks and found competitive performance.
\end{abstract}


\section{Introduction}
\label{sec:intro}
Sensors are heavily used in a wide number of time-series, industrial and measurement problems, such as for energy management, industrial IoT, smart cities, agriculture, healthcare, smart homes, fraud detection, autonomous systems etc. However, the reliability of the models (based on these sensor inputs) or the predictions of these models is only as good as the reliability of data from the sensors. Many a times, due to faults, the sensors do not transmit data. In such scenarios, the input features to a model may vary, leading to the problem of variation in input size. Similarly, in data collection under a multi-entity environment, the failure of some actors may directly influence the model's input feature space. These challenging scenarios where the input space does not stay fixed, can be categorized under a single term called \textit{haphazard inputs} \cite{AuxDrop2023agarwal}. This category of problems have not been typically studied until recent past, where Agarwal \textit{et al.} have tried to formalize the problem domain, identifying important shared characteristics among problems \cite{AuxDrop2023agarwal,AuxNet2022agarwal}. Making better models to address haphazard inputs is an important area of research and bears significant economic value as well.

The typical machine learning models assume that the dimensions of the input space are fixed and do not change during training or inference. However, haphazard inputs, challenge this assumption and force us to rethink about typical machine learning models. This is a very new field and a major contribution has been made by Agarwal \textit{et al.} \cite{AuxDrop2023agarwal,AuxNet2022agarwal}. Their idea works by making per input models that interact with each other based on the hedging algorithm \cite{odl2018sahoo}. Hedging regulates the weight/contribution of each input towards model's final prediction. However, we have identified that hedging based approaches have multiple disadvantages like, they require to make models for each input, which is challenging to generalize and scale to other modalities such as images or videos, they require online gradient descent algorithm which adds to the complexity, they require more number of hyperparameters which makes them hyperparameter-sensitive, moreover, they are harder to implement and highly complex leading to limited ease of access for different applications.

In this work, we propose to model this problem with a simpler approach based on self-attention \cite{vaswani2017attention}. We first show that hedging can be approximated as weighted residual network. Using this intuition we design a self-attention based model called HapNet, which performs competitively with the current state-of-the-art models, however is quite straight forward to implement and thus have better generalizability and scalability. We study the haphazard inputs problem in detail and propose an even challenging and realistic version of the problem where the inputs are positionally uncorrelated. To address this challenging case, we extend the proposed HapNet model to HapNetPU model, which incorporates the idea of feed-back loop. In order to demonstrate the effectiveness of the proposed approach, we have used 5 benchmark datasets from different time-series problem domains. \textbf{Novelty.} In the spirit of Occam's razor principle, the novelty of our work lies in providing a simpler yet effective approach to solve haphazard inputs based problems in both positionally correlated and uncorrelated scenarios.

\textbf{Contributions.}
\textbf{(a)} Unlike AuxDrop or the other hedging based methods \cite{odl2018sahoo,AuxNet2022agarwal}, the proposed method does not need to build per feature model. It uses joint optimization which is a more general approach and may lead to enhancement in sharing and mutual learning of feature space for general problems of different modalities and larger input sizes like images or videos \cite{gato2022reedgeneralist,byol2020}.
\textbf{(b)} The proposed approach uses simple self-attention and therefore is free from online gradient descent based backpropagation. Furthermore, our work also contributes by showing that for one dimensional inputs, the embedding layer is not required in transformers, and the input can directly serve as its own embedding. This is in contrast to the previous findings \cite{multibit2023born}.
\textbf{(c)} We contribute a more challenging and realistic case of haphazard inputs, where the features become more positionally uncorrelated than the typical case. We also propose HapNetPU model to tackle this challenging scenario. Note that HapNetPU is the only model that could work for this kind of problems.
\textbf{(d)} We compare the proposed models with state-of-the-art models and perform extensive ablation study.

\textbf{Related work.}
Earlier works on online machine learning utilized basic machine learning models like $k$ nearest neighbors \cite{knn2006}, decision trees \cite{decisiontrees2000}, support vector machines \cite{svm2007}, fuzzy models \cite{fuzzy2010}, neural networks \cite{neuralnets2013} etc. Recent methods based on deep learning, like \cite{odl2018sahoo}, have also shown competitive performance for online machine learning. However, the main challenge with these methods remain that they cannot be directly employed for non-fixed input space problems. Some other methods based on incremental learning \cite{incremental2012} or online learning \cite{olvf2019}, have also attempted to address these problems, however, they are not based on deep learning. In this direction, important contributions have been made by Agarwal \textit{et al.} \cite{AuxDrop2023agarwal,AuxNet2022agarwal}. These works mainly rely on a learning mechanism called ``hedging'' \cite{odl2018sahoo}. Our work builds upon these hedging based methods.

\begin{figure}[t]
    \centering
    \begin{subfigure}[b]{0.8\textwidth}
        \centering
        \includegraphics[width=\textwidth]{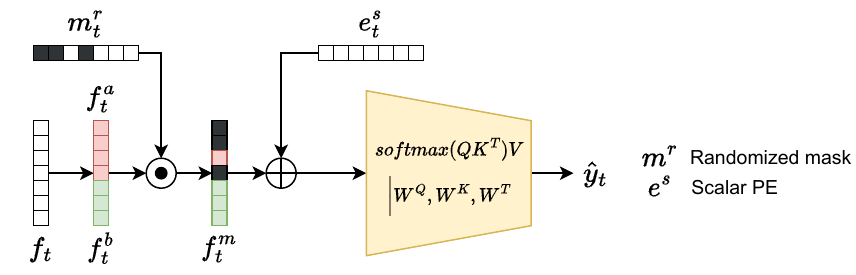}
        \caption{HapNet}
        \label{subfig:HapNet}
    \end{subfigure}
    \hspace{2em}
    \begin{subfigure}[b]{0.8\textwidth}
        \centering
        \includegraphics[width=\textwidth]{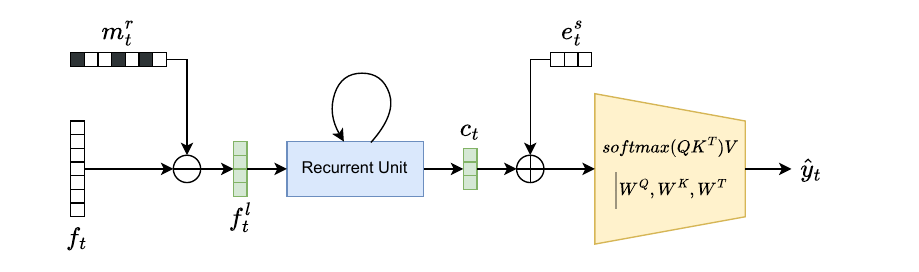}
        \caption{HapNetPU}
        \label{subfig:HapNetPU}
    \end{subfigure}
    \caption{The two proposed models are shown in (a) HapNet (for positionally correlated), (b) HapNetPU (for positionally uncorrelated case). Both explain how an input feature $f_t$ is processed. $\bigodot$ implies Hadamard product, $\circleddash$ implies inverse Hadamard, the operation of removing the values from specific positions and compressing the dimensions of the input feature at time $t$ in order to make them positionally uncorrelated. 
    $f_t$ is input features at time $t$,
    $f^b_t$ is base features at time $t$,
    $f^a_t$ is auxiliary features at time $t$,
    $f^m_t$ is masked features at time $t$,
    $m^r_t$ is randomized mask at time $t$,
    $e^s_t$ is scalar position encoding at time $t$,
    $f^l_t$ is remaining features after feature loss at time $t$,
    $c_t$ is context at time $t$,
    $\hat{y}_t$ is predicted label at time $t$.
    }
    \label{fig:model}
\end{figure}

\section{Proposed Approach}
We use a self-attention based mechanism to capture the correlations among the input features. During training, each input feature at time $t$ is randomly masked and used for training. The proposed approach is straight forward as shown in Fig. \ref{subfig:HapNet}. We first formally explain the haphazard input problem. Then the role of hedging is discussed, followed by the proposed approach.

\textbf{Task formulation.}
Assuming a faulty data source or a sensor, generating the input data stream $\mathcal{D}$, the objective of online learning with haphazard inputs is to correctly label the sample $f_t \in \mathcal{D}$ arriving at time step $t$, and simultaneously update the model parameters with the gradient of prediction loss at $t$. We assume that since the data source is faulty, a subset of the sample $f_t$ is called auxiliary, denoted by $f^a_t$, and contains reliable information with a probability of availability of data, $p$. When $p=1$, all information in $f^a_t$ is available and fully reliable. The difference or base subset of $f_t$, denoted by $f^b_t$, is mutually exclusive to $f^a_t$ and is calculated as the difference between $f_t$ and $f^a_t$. Unlike $f^a_t$, the data in $f^b_t$ is always available and reliable. The online model works in a predict-then-update manner where the prediction for $f_{t+1}$ cannot be made before the online model is updated with prediction for $f_t$. In other words, the time-series data offers a feature to be learned. Each such feature consists of reliable (base) and unreliable (auxiliary) features. The model is tasked with online learning with this time-series data.

\textbf{Hedging as weighted residuals.}
Hedging algorithm proposed by~\cite{odl2018sahoo} has severed as the primary driving force of many haphazard data processing methods. The main idea is to create an ensemble of classifiers where the outcome of each classifier is weighted by a scalar value, as shown below:
\begin{equation}
    \hat{y}^E_t = \sum \alpha_i\, \hat{y}^C_{t,i},
\end{equation}
\begin{equation}\label{eq:odl2}
    \hat{y}^C_{t,i} = \sigma(\Theta_i\, \sigma(W_i\, h_{i-1})),
\end{equation}
where, $\hat{y}^E_t$ is the ensemble's label prediction, $\hat{y}^C_{t,i}$ is the layer's label prediction of the $i^{th}$ classifier, $\Theta_i$ is the classification weight matrix of the $i^{th}$ classifier, $\alpha_i$ is classifier weight, $\sigma$ is the activation function, $W_i$ is the layer's weight matrix, and $h_{i-1}$ is the activation from the previous layer. In hedging, a label prediction loss is imposed on every classifier's prediction $\hat{y}^C_{t,i}$ and also on the ensemble's prediction $\hat{y}^C_t$. The output of a fully residual network (like densenet), without an ensemble loss unlike Eq.~\eqref{eq:odl2}, can also be written as:
\begin{equation}\label{eq:weightedresidual}
    \begin{split}
        \hat{y}^R_t &= \sum \alpha^\prime_i\, \sigma(\Theta_i^\prime\, \sigma(W^\prime_i\, h^\prime_{i-1})) \\
        &= \sum \alpha^\prime_i\, \sigma(W^{\prime\prime}_i\, h^\prime_{i-1}).
    \end{split}
\end{equation}
where, $W^{\prime\prime}_i$ approximates the compound layer operation. In Eq.~\eqref{eq:weightedresidual}, if $\alpha^\prime_i=1$, the model behaves like a vanilla residual network, and if  $\alpha^\prime_i\neq\{1,0\}$, then the model becomes a weighted residual network. This architectural similarity between hedging and weighted residual approach indicates that hedging uses coarse-level attention. Inspired by this similarity, we propose to replace this approach with a simple yet effective, fine intra attention approach.

\textbf{HapNet.}
The proposed HapNet model (Fig. \ref{subfig:HapNet}) is based on self-attention. Each feature $f_t$ is randomized/bootstrapped on its auxiliary features to produce multiple masked versions of $f^a_t$ which leads to multiple copies of $f_t$. These copies are directly passed to the positional embedding layer $e^s_t$, that adds the positional information. Note that the feature values do not pass through any embedding layer themselves and are directly used as their corresponding embeddings. This was experimentally discovered by us that it is better to pass the values directly. This discovery is in direct contrast to \cite{multibit2023born}. The resulting vector is passed through an encoder based on self-attention. In our specific model we have used transformers to model that. At the inference time, the randomization module on $f^a_t$ is turned off, so that the feature $f_t$ directly passes through the encoder. We have used cross-entropy loss for training of HapNet.

\textbf{HapNetPU.}
We discovered that although haphazard inputs pose significant challenge to typical deep learning models, however, a more realistic situation could be when the inputs are completely unavailable and in those scenarios it is hard to match their unavailability with their positions. Assumption is that the order of features is maintained. Under such scenarios, no existing models could work. To tackle this problem, we present HapNetPU (Fig. \ref{subfig:HapNetPU}. HapNetPU builds upon HapNet by employing a feed-back operator (can be realized by any recurrent network like LSTM or GRU). The resulting feature $f^l_t$ loses its original dimension and is truly haphazard in behavior. HapNetPU, due to its recurrence operation, is able to process it though.

\section{Experiments}
This section presents the experimental protocol and details, the dataset used, results, and analysis. We use state-of-the-art models for processing haphazard inputs \cite{AuxDrop2023agarwal,AuxNet2022agarwal} and other deep learning methods \cite{odl2018sahoo,olvf2019} for comparison to the proposed method. Average error i.e. number of incorrect predictions out of total samples tested is used as the evaluation metric. Along with that, the cross entropy (CE) loss is used to train and test the models. Rarely, macro and micro F1 scores and accuracy have been used as alternative metrics. The encoder consists of a transformer with 6 blocks, a dropout of .15 is used with layer normalization and two liner layers on the outputs of self-attention. If not reported otherwise, the learning rate was 0.0001 with Adam optimizer. All experiments were repeated 20 times and their mean and standard deviations have been reported.

\textbf{Datasets}
Five datasets have been used in our experiments. These are (a) Italy power demand dataset \cite{italypower} (b) \textit{german}, \textit{svmguide3}, \textit{magic04}, \textit{a8a} \cite{uci2007}. All these datasets contain time-series data from different domains and applications. More details can be found in Agarwal \textit{et al.} \cite{AuxDrop2023agarwal}.

\begin{table}[t!]
\centering
\caption{The CE loss on Italy power demand dataset are shown for AuxNet, AuxDrop (ODL), HapNet, along with errors for HapNet. Best results are marked as \textbf{bold} and the {\ul second best} are underlined at each probability value. Each HapNet experiment was repeated 20 times. Probability values corresponds to the availability of auxiliary features.}
\label{tab:table3 italy power demand}
\begin{tabular}{lcccc}
\toprule
Probability & AuxNet & AuxDrop             & HapNet                 & HapNet (Error) \\ \midrule
0.5         & 0.6975 & {\ul 0.6031±0.0081} & \textbf{0.4512±0.0187} & 234.92±17.36   \\
0.6         & 0.6831 & {\ul 0.5839±0.0111} & \textbf{0.4092±0.0146} & 204.30±11.35   \\
0.7         & 0.6788 & {\ul 0.5497±0.0082} & \textbf{0.3726±0.0231} & 179.14±14.29   \\
0.8         & 0.6130 & {\ul 0.5321±0.0071} & \textbf{0.3203±0.0127} & 146.10±7.66    \\
0.9         & 0.5456 & {\ul 0.5149±0.0119} & \textbf{0.2780±0.0151} & 119.50±7.26    \\
0.95        & 0.5168 & {\ul 0.5013±0.0108} & \textbf{0.2462±0.0152} & 100.78±9.63    \\
0.99        & 0.5165 & {\ul 0.4788±0.0101} & \textbf{0.2254±0.0158} & 89.26±10.12    \\ \bottomrule
\end{tabular}%
\end{table}

\begin{table}[t!]
\centering
\caption{Comparison of error among OLVF, AuxDrop (ODL), AuxDrop (OGD), HapNet is shown for different datasets. Best results are marked as \textbf{bold} and the {\ul second best} are underlined at each probability value. The error is reported as the mean ± standard deviation of the 20 experiments performed with random seeds. Probability values corresponds to the availability of auxiliary features.}
\label{tab:table4}
\begin{tabular}{lccccc}
\toprule
Probability & Dataset   & OLVF        & AuxDrop (ODL)      & AuxDrop (OGD)        & HapNet               \\ \midrule
0.73        & \textit{german}    & 333.4±9.7   & \textbf{300.4±4.4} & 312.8±19.3           & {\ul 307.6±1.6}      \\
0.72        & \textit{svmguide3} & 346.4±11.6  & \textbf{297.2±2.0} & {\ul 297.5±1.5}      & 299.8±0.2            \\
0.68        & \textit{magic04}   & 6152.4±54.7 & {\ul 5536.7±59.3}  & \textbf{5382.8±98.9} & 6391.7±124.1         \\
0.75        & \textit{a8a}       & 8993.8±40.3 & {\ul 6710.7±117.8} & 7313.5±277.7         & \textbf{6514.4±87.6} \\ \bottomrule
\end{tabular}%
\end{table}

\begin{table}[t!]
\centering
\caption{Comparison of average number of errors among OLSF, OLVF, AuxDrop (ODL), AuxDrop (OGD), HapNet is shown on trapezoidal data streams. Best results are marked as \textbf{bold} and the {\ul second best} are underlined at each probability value. The error is reported as the mean ± standard deviation of the 20 experiments performed with random seeds.}
\label{tab:table5}
\begin{tabular}{lccccc}
\toprule
Dataset   & \multicolumn{1}{l}{OLSF} & OLVF              & AuxDrop (ODL)   & AuxDrop (OGD)         & HapNet               \\ \midrule
\textit{german}    & 385.5±10.2               & 329.2±9.8         & {\ul 312.2±8.0} & 320.9±39.4            & \textbf{301.2±1.9}   \\
\textit{svmguide3} & 361.7±29.7               & 351.6±25.9        & {\ul 296.9±1.0} & 297.0±0.9             & \textbf{295.7±8.3}   \\
\textit{magic04}   & 6147.4±65.3              & {\ul 5784.0±52.7} & 6361.25±319.6   & \textbf{5635.8±100.9} & 6671.5±98.1          \\
\textit{a8a}       & 9420.4±549.9             & 8649.8±526.7      & 7850.9±15.9     & {\ul 7848.8±10.3}     & \textbf{7831.5±21.0} \\ \bottomrule
\end{tabular}%
\end{table}

\begin{table}[t!]
\centering
\caption{Comparison in terms of the average error and CE loss is shown for the proposed HapNetPU on three datasets. Each experiment was repeated 20 times.}
\label{tab:HapNetPU}
\begin{tabular}{lccc}
\toprule
Dataset     & Probability & Error     & Loss          \\ \midrule
\textit{Italy power} & 0.8         & 512.7±8.8 & 0.7086±0.0035 \\
\textit{german}      & 0.73        & 326.4±0.4 & 0.6470±0.0020 \\
\textit{svmguide3}   & 0.72        & 301.0±1.4 & 0.5695±0.0022 \\ \bottomrule
\end{tabular}%
\end{table}

\begin{table}[t!]
\centering
\caption{In this ablation the dropout scores are varied on the \textit{german} dataset to see its effect on the performance of the HapNet method, both in terms of the average error and CE loss.}
\label{tab:ablation dropout}
\begin{tabular}{lcccc}
\toprule
Data stream & Dropout & Probability & Error     & CE Loss       \\ \midrule
Trapezoid   & 0.5     & 0.73        & 322.8±3.4 & 0.6433±0.0030 \\
Trapezoid   & 0.3     & 0.73        & 305.4±8.7 & 0.6241±0.0065 \\
Haphazard   & 0.5     & 0.73        & 325.6±0.3 & 0.6389±0.0023 \\
Haphazard   & 0.3     & 0.73        & 317.2±1.3 & 0.6305±0.0011 \\ \bottomrule
\end{tabular}%
\end{table}

\begin{table}[t!]
\centering
\caption{In this ablation the probability of presence of auxiliary inputs is varied on the \textit{german} dataset to see its effect on the performance of the HapNet method, both in terms of the average error and CE loss.}
\label{tab:ablation probability}
\begin{tabular}{lcccc}
\toprule
Probability & 0.6    & 0.7    & 0.8    & 0.9    \\ \midrule
Loss        & 0.6268 & 0.6209 & 0.6193 & 0.6091 \\
Error       & 310.1  & 308.4  & 303.9  & 294.4  \\
Macro F1    & 0.4772 & 0.4841 & 0.4959 & 0.5101 \\
Micro F1    & 0.6898 & 0.6915 & 0.6960 & 0.7055 \\
Accuracy    & 0.6898 & 0.6915 & 0.6960 & 0.7055 \\ \bottomrule
\end{tabular}%
\end{table}

\begin{table}[t!]
\centering
\caption{In this ablation the learning rate is varied on the \textit{german} and \textit{svmguide3} datasets with six encoders and batchsize 64 to see its effect on the performance of the HapNet method, both in terms of the average error and CE loss.}
\label{tab:ablation learning rate}
\begin{tabular}{lccc}
\toprule
Learning rate & Dataset                       & Error                     & Loss                       \\ \midrule
0.001         & \textit{german}                        & 325.2                     & 0.6473                     \\
0.001         & \textit{german}                        & 319.9                     & 0.6403                     \\
0.00001       & \textit{german}                        & 316.4                     & 0.6219                     \\
0.001         & \multicolumn{1}{l}{svmguide3} & \multicolumn{1}{l}{303.6} & \multicolumn{1}{l}{0.5762} \\
0.001         & \textit{svmguide3}                     & 300.8                     & 0.5657                     \\
0.00001       & \textit{svmguide3}                     & 299.3                     & 0.5552                     \\ \bottomrule
\end{tabular}%
\end{table}

\begin{table}[t!]
\centering
\caption{In this ablation the number of encoders are varied on the \textit{german} and \textit{svmguide3} datasets with learning rate 0.0001 and batchsize 64 to see its effect on the performance of the HapNet method, both in terms of the average error and CE loss.}
\label{tab:ablation encoders}
\begin{tabular}{lccc}
\toprule
Number of encoders & Dataset   & Error & Loss   \\ \midrule
12                 & \textit{german}    & 306.5 & 0.6205 \\
24                 & \textit{german}    & 306.6 & 0.6197 \\
12                 & \textit{svmguide3} & 301.2 & 0.5615 \\
24                 & \textit{svmguide3} & 301.7 & 0.5616 \\ \bottomrule
\end{tabular}%
\end{table}

\begin{table}[h!]
\centering
\caption{In this ablation the batchsize is varied on the \textit{german} and \textit{svmguide3} datasets with learning rate 0.0001 and 6 number of encoders to see its effect on the performance of the HapNet method, both in terms of the average error and CE loss.}
\label{tab:ablation batchsize}
\begin{tabular}{lccc}
\toprule
Batch size & Dataset   & Error & Loss   \\ \midrule
16         & \textit{german}    & 306.9 & 0.6193 \\
32         & \textit{german}    & 309.8 & 0.6211 \\
128        & \textit{german}    & 306.1 & 0.6184 \\
16         & \textit{svmguide3} & 300.1 & 0.5605 \\
32         & \textit{svmguide3} & 301.6 & 0.5616 \\
128        & \textit{svmguide3} & 301.1 & 0.5619 \\ \bottomrule
\end{tabular}%
\end{table}

\textbf{Results and ablation.}
We performed several experiments with different types of datasets Table \ref{tab:table3 italy power demand}, \ref{tab:table4}, \ref{tab:table5} to evaluate the performance of the proposed method. The proposed approach has shown consistent performance across different datasets, metrics, and evaluation settings. Note that Table \ref{tab:table5} uses trapezoid inputs \cite{AuxDrop2023agarwal}. It is particularly worth noting in Table \ref{tab:table3 italy power demand}, HapNet outperforms AuxNet, approximately, by a factor of two. Table \ref{tab:HapNetPU} shows the results for the positionally non-correlated setting. Note that the performance of HapNetPU is not lagging significantly on \textit{german} and \textit{svmguide3} datasets. This shows strong capability of HapNetPU to model even under positionally non-correlated haphazard inputs.
Further, we have performed several ablations on the role of dropout, suitability of number of encoders, batchsizes, learning rates and the variation of model performance with changing probability. These can be referenced in Table \ref{tab:ablation batchsize}, \ref{tab:ablation dropout}, \ref{tab:ablation encoders}, \ref{tab:ablation learning rate}, \ref{tab:ablation probability}.

\subsection{Conclusion}
This work propose a simple yet effective algorithm for haphazard input based problems. It shows that the more widely used hedging based algorithm can be well estimated by a more generic version -- self-attention, leading to better generalization, scalability, ease of use and adaptability to different modalities. The proposed methods achieved competitive performance to other state-of-the-art methods. We further introduced a more challenging case of haphazard inputs where the features values are positionally non-correlated. We showed that a feed-back based modification of the proposed HapNet may address this challenging case as well. Extensive ablations on the five datasets revealed the effectiveness of the proposed HapNet approach. However there are some limitations to the proposed work.
\textbf{Limitation.}
Although the HapNet approach shows competitive performance, however, in some cases the other methods outperform it. It shows that a more optimized variant of HapNet can be developed. In place of LSTMs in HapNetPU, some other network like GRU or transformers may be used to further improve its performance.
\textbf{Future scope.}
Our approach can be extended to images and videos as the transformer architecture is omnipresent and the data can be easily divided. Different techniques may be used to replace the feed-back module in HapNetPU to further improve its performance.



\begin{thebibliography}{10}

\bibitem{AuxDrop2023agarwal}
Rohit Agarwal, Deepak Gupta, Alexander Horsch, and Dilip~K Prasad.
\newblock Aux-drop: Handling haphazard inputs in online learning using auxiliary dropouts.
\newblock {\em arXiv preprint arXiv:2303.05155}, 2023.

\bibitem{AuxNet2022agarwal}
Rohit Agarwal, Krishna Agarwal, Alexander Horsch, and Dilip~K Prasad.
\newblock Auxiliary network: Scalable and agile online learning for dynamic system with inconsistently available inputs.
\newblock In {\em International Conference on Neural Information Processing}, pages 549--561. Springer, 2022.

\bibitem{odl2018sahoo}
Doyen Sahoo, Quang Pham, Jing Lu, and Steven~CH Hoi.
\newblock Online deep learning: learning deep neural networks on the fly.
\newblock In {\em Proceedings of the 27th International Joint Conference on Artificial Intelligence}, pages 2660--2666, 2018.

\bibitem{vaswani2017attention}
A~Vaswani.
\newblock Attention is all you need.
\newblock {\em Advances in Neural Information Processing Systems}, 2017.

\bibitem{gato2022reedgeneralist}
Scott Reed, Konrad Zolna, Emilio Parisotto, Sergio~G{\'o}mez Colmenarejo, Alexander Novikov, Gabriel Barth-maron, Mai Gim{\'e}nez, Yury Sulsky, Jackie Kay, Jost~Tobias Springenberg, et~al.
\newblock A generalist agent.
\newblock {\em Transactions on Machine Learning Research}, 2020.

\bibitem{byol2020}
Jean-Bastien Grill, Florian Strub, Florent Altch{\'e}, Corentin Tallec, Pierre Richemond, Elena Buchatskaya, Carl Doersch, Bernardo Avila~Pires, Zhaohan Guo, Mohammad Gheshlaghi~Azar, et~al.
\newblock Bootstrap your own latent-a new approach to self-supervised learning.
\newblock {\em Advances in neural information processing systems}, 33:21271--21284, 2020.

\bibitem{multibit2023born}
Jannis Born and Matteo Manica.
\newblock Regression transformer enables concurrent sequence regression and generation for molecular language modelling.
\newblock {\em Nature Machine Intelligence}, 5(4):432--444, 2023.

\bibitem{knn2006}
Charu~C Aggarwal, Jiawei Han, Jianyong Wang, and Philip~S Yu.
\newblock A framework for on-demand classification of evolving data streams.
\newblock {\em IEEE Transactions on Knowledge and Data Engineering}, 18(5):577--589, 2006.

\bibitem{decisiontrees2000}
Pedro Domingos and Geoff Hulten.
\newblock Mining high-speed data streams.
\newblock In {\em Proceedings of the sixth ACM SIGKDD international conference on Knowledge discovery and data mining}, pages 71--80, 2000.

\bibitem{svm2007}
Ivor~W Tsang, Andras Kocsor, and James~T Kwok.
\newblock Simpler core vector machines with enclosing balls.
\newblock In {\em Proceedings of the 24th international conference on Machine learning}, pages 911--918, 2007.

\bibitem{fuzzy2010}
Tor Das et~al.
\newblock A novel incremental rough set-based pseudo outer-product with ensemble learning (ierspop) neuro-fuzzy system for forecasting volatility.
\newblock 2010.

\bibitem{neuralnets2013}
Daniel Leite, Pyramo Costa, and Fernando Gomide.
\newblock Evolving granular neural networks from fuzzy data streams.
\newblock {\em Neural Networks}, 38:1--16, 2013.

\bibitem{incremental2012}
Robi Polikar.
\newblock Ensemble learning.
\newblock {\em Ensemble machine learning: Methods and applications}, pages 1--34, 2012.

\bibitem{olvf2019}
Ege Beyazit, Jeevithan Alagurajah, and Xindong Wu.
\newblock Online learning from data streams with varying feature spaces.
\newblock In {\em Proceedings of the AAAI conference on artificial intelligence}, volume~33, pages 3232--3239, 2019.

\bibitem{italypower}
Hoang~Anh Dau, Anthony Bagnall, Kaveh Kamgar, Chin-Chia~Michael Yeh, Yan Zhu, Shaghayegh Gharghabi, Chotirat~Ann Ratanamahatana, and Eamonn Keogh.
\newblock The ucr time series archive.
\newblock {\em IEEE/CAA Journal of Automatica Sinica}, 6(6):1293--1305, 2019.

\bibitem{uci2007}
Arthur Asuncion, David Newman, et~al.
\newblock Uci machine learning repository, 2007.

\end{thebibliography}

\end{document}